\documentclass[letterpaper]{article} % DO NOT CHANGE THIS
\usepackage{aaai2026}  % DO NOT CHANGE THIS
\usepackage{times}  % DO NOT CHANGE THIS
\usepackage{helvet}  % DO NOT CHANGE THIS
\usepackage{courier}  % DO NOT CHANGE THIS
\usepackage[hyphens]{url}  % DO NOT CHANGE THIS
\usepackage{graphicx} % DO NOT CHANGE THIS
\usepackage{natbib}  % DO NOT CHANGE THIS AND DO NOT ADD ANY OPTIONS TO IT
\usepackage{caption} % DO NOT CHANGE THIS AND DO NOT ADD ANY OPTIONS TO IT
\usepackage{algorithm}
\usepackage{algorithmic}

\usepackage{newfloat}
\usepackage{listings}
\DeclareCaptionStyle{ruled}{labelfont=normalfont,labelsep=colon,strut=off} % DO NOT CHANGE THIS
\floatstyle{ruled}
\newfloat{listing}{tb}{lst}{}
\floatname{listing}{Listing}
\usepackage{multirow}
\usepackage{booktabs}

\title{HalluClean: A Unified Framework to Combat Hallucinations in LLMs}

\author{
    Yaxin Zhao\textsuperscript{\rm 1},
    Yu Zhang\textsuperscript{\rm 1}
}
\affiliations{
    \textsuperscript{\rm 1}Harbin Institute of Technology, Harbin, China\\
    \{yxzhao, zhangyu\}@ir.hit.edu.cn
}

\usepackage{bibentry}
\begin{document}

\maketitle

\begin{abstract}
Large language models (LLMs) have achieved impressive performance across a wide range of natural language processing tasks, yet they often produce hallucinated content that undermines factual reliability. To address this challenge, we introduce \textbf{HalluClean}, a lightweight and task-agnostic framework for detecting and correcting hallucinations in LLM-generated text. HalluClean adopts a \textbf{reasoning-enhanced paradigm}, explicitly decomposing the process into planning, execution, and revision stages to identify and refine unsupported claims. It employs \textbf{minimal task-routing prompts} to enable \textbf{zero-shot generalization} across diverse domains, without relying on external knowledge sources or supervised detectors. We conduct extensive evaluations on five representative tasks—question answering, dialogue, summarization, math word problems, and contradiction detection. Experimental results show that HalluClean significantly improves factual consistency and outperforms competitive baselines, demonstrating its potential to enhance the trustworthiness of LLM outputs in real-world applications.
\end{abstract}
% Uncomment the following to link to your code, datasets, an extended version or similar.
% You must keep this block between (not within) the abstract and the main body of the paper.
%  \begin{links}
%     \link{Code}{https://github.com/tom68-ll/SAFENLIDB}
%     % \link{Datasets}{https://aaai.org/example/datasets}
%     \link{Extended version}{https://arxiv.org/abs/2511.06778}
% \end{links}

\section{Introduction}

Large language models (LLMs) have revolutionized natural language processing (NLP), powering applications such as conversational agents, content generation, and decision support systems~\citep{chowdhery2023palm,touvron2023llama,bang2023multitask,qin2023chatgpt}. These models leverage large-scale pretraining and are further enhanced via instruction tuning~\citep{chung2024scaling,wang2022super,wang2022self} and alignment techniques that optimize for human preferences~\citep{ouyang2022training,achiam2023gpt}. However, despite their remarkable fluency and versatility, LLMs frequently produce hallucinated or factually incorrect content~\citep{huang2023survey,ji2023survey}, undermining their reliability in safety-critical contexts.

\begin{figure*}[ht]
\centering
\includegraphics[width=\textwidth]{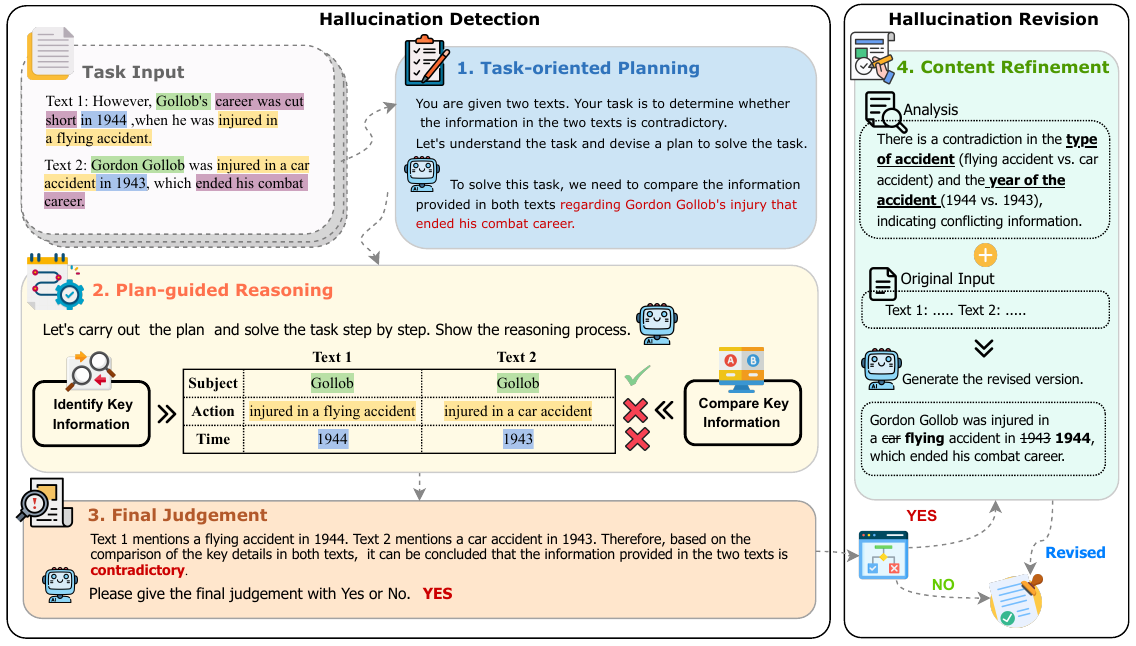}
\caption{ Overview of the HalluClean framework. It consists of two modules: hallucination detection and revision. The detection module generates a task-specific plan, performs step-by-step reasoning, and makes a final judgment. If a hallucination is detected, the revision module revises the content based on the identified reasoning to eliminate hallucinated information.}
\label{fig:framework}
\end{figure*}

Existing hallucination mitigation approaches typically fall into two categories. Retrieval-augmented generation methods~\citep{varshney2023stitch,cao2023step,kang2023ever,rawte2023troubling} query external knowledge sources to validate or correct model outputs. Meanwhile, supervised detection approaches~\citep{razumovskaia2024dial,zhang2023r,qiu2023think} rely on human-labeled data to train classifiers that identify hallucinations. While both strategies have shown promise, they suffer from key limitations: retrieval-based methods depend on the availability and accuracy of external sources, and annotation-based methods are costly and poorly generalize to novel hallucination types. Furthermore, hallucination behaviors vary widely across tasks—such as question answering~\citep{zheng2023doeschatgptfallshort}, summarization~\citep{cao2022hallucinated}, and dialogue~\citep{das2022diving}—yet most prior work focuses narrowly on specific settings~\citep{mundler2023self}, limiting scalability and robustness.

To address these challenges, we propose a lightweight, task-agnostic framework for hallucination detection and correction that operates without external knowledge or task-specific supervision. Our approach leverages minimal task descriptions to instantiate a task-adaptive interface, guiding LLMs through a reasoning-enhanced, zero-shot process. Inspired by the plan-and-solve paradigm, we decompose hallucination mitigation into explicit planning and execution phases, enabling models to locate unsupported claims and revise outputs with improved factuality.

We introduce \textbf{HalluClean}, a unified framework that detects and corrects hallucinations in LLM-generated outputs via structured reasoning. HalluClean employs compact prompts to elicit multi-step reasoning traces, which serve both to identify hallucinated segments and to guide targeted revision. This plug-and-play, prompt-based design ensures broad applicability across model architectures and NLP tasks, while remaining compatible with open-source LLMs for privacy-preserving deployments.

\vspace{0.5em}
\noindent Our contributions are summarized as follows:
\begin{itemize}
  \item \textbf{HalluClean Framework}: We present a zero-shot hallucination detection and correction framework based on structured reasoning. HalluClean is modular and supports flexible integration with diverse LLMs, including open-source models.
  \item \textbf{Task-Agnostic Generalization}: HalluClean achieves strong performance across a variety of NLP tasks—question answering, summarization, dialogue, math word problems, and contradiction detection—without requiring task-specific fine-tuning.
  \item \textbf{Domain-Level Robustness}: We demonstrate HalluClean's effectiveness in domain-sensitive settings such as medicine and finance, highlighting its potential for deployment in real-world, high-stakes applications.
\end{itemize}

\section{Related Work}

\subsection{Hallucinations in LLMs}Hallucinations in large language models (LLMs) have been extensively studied, focusing on their causes~\citep{pan2023risk,chen2024combating,kasai2024realtime,wang2023causal,lee2022factuality,yao2023editing}, evaluation methodologies~\citep{lin2021truthfulqa,lee2022factuality,min2023factscore,li2023halueval}, and behavioral analysis~\citep{zhao2023knowing,dong2024statistical,li2023benchmarking}. Many studies have investigated ways to mitigate hallucinations through retrieval-augmented generation~\citep{peng2023check,varshney2023stitch,kang2023ever} and supervised fine-tuning~\citep{elaraby2023halo,razumovskaia2024dial,zhang2023r}. To improve LLM reliability, researchers have explored prompting-based solutions. For example, \citet{si2022prompting} proposed simple yet effective prompts that enhance GPT-3’s factual accuracy, while \citet{mitchell2022enhancing} introduced a two-model framework in which one model generates responses and another evaluates their logical coherence. More recently, \citet{mundler2023self} found that 17.7\% of ChatGPT-generated sentences contain self-contradictions and proposed a three-step pipeline to detect and mitigate them without relying on external knowledge. Despite these advancements, hallucination detection and mitigation in LLMs remain challenging. Our method uses task-adaptive prompts in a zero-shot setting to guide LLMs in detecting and revising hallucinated content by eliciting and leveraging model reasoning.

\begin{table*}[t]
\centering
\begin{tabular}{p{3 cm}|p{12cm}}
\toprule
\textbf{Task Type} & \textbf{Task Routing Prompt} \\
\midrule
Question Answering & You are provided with a question and its corresponding answer. Your task is to determine whether the answer contains hallucinated content.  \\
\midrule
Dialogue Systems & You are provided with a dialogue history and its corresponding response.
Your task is to determine whether the response contains hallucinated content.  \\
\midrule
Summarization & You are provided with a document and its corresponding summary.
Your task is to determine whether the summary contains hallucinated content. \\
\midrule
Math Word Problems & You are provided with a math word problem.
Your task is to determine whether the problem is unanswerable. \\
\midrule
Self-contradiction & You are given two texts. 
Your task is to determine whether the information in the two texts is contradictory.  \\
\bottomrule
\end{tabular}
\caption{Task-oriented routing prompts for different NLP applications. These concise instructions guide the model to understand the specific hallucination detection objective for each task type.}
\label{tab:task-prompts}
\end{table*}

\subsection{Advancements in Prompting Techniques}Prompting strategies have played a crucial role in enhancing the reasoning abilities of LLMs. Chain-of-Thought (CoT) prompting~\citep{wei2022chain} explicitly structures intermediate reasoning steps, significantly improving model performance on complex reasoning tasks. Building upon this, various enhancements have been proposed, including prompt ensembling~\citep{wang2022self,li2022advance,fu2022complexity}, problem decomposition~\citep{zhou2022least,khot2022decomposed,dua2022successive}, and structured planning methods~\citep{yao2022react,huang2022language,wang2023describe,liu2023llm+}. To reduce manual effort and computational overhead, zero-shot CoT prompting~\citep{kojima2022large} was introduced, allowing LLMs to autonomously generate reasoning steps without the need for labeled exemplars. However, existing methods primarily focus on general reasoning tasks, and limited work has explored their applicability in addressing hallucinations within LLM-generated text.

In this work, we propose a new prompt-based mechanism to enhance the effectiveness of hallucination detection and ensuring more reliable correction.

\section{Method}\label{method}
We introduce \textbf{HalluClean}, a task-agnostic framework for hallucination detection and correction in LLMs. It leverages structured reasoning guided by minimal task prompts and operates in zero-shot settings.

\subsection{Task-Based Categorization of LLM Hallucinations}
LLMs support diverse applications—summarization, QA, dialogue, and problem solving—but frequently generate factually inconsistent or logically contradictory content, known as \textit{hallucinations}. These typically arise when generated outputs lack grounding in verifiable knowledge or logic. To address this, we adopt a task-based categorization of hallucinations across five representative NLP scenarios:

\paragraph{Question Answering} Hallucinations manifest as unsupported claims, misinterpreted context, or factual errors that deviate from the input or common knowledge.

\paragraph{Dialogue Systems} Errors often stem from entity mismatches—substituting similar, dissimilar, or cross-type entities—leading to factual incoherence with the dialogue history.

\paragraph{Summarization} Generated summaries may include unverifiable details or fabricate facts not grounded in the source text, often misrepresenting entities or relations.

\paragraph{Math Word Problems} Under-specified or ill-posed problems cause hallucinations when essential constraints are missing, variables are vague, or assumptions violate logic (e.g., negative quantities where not allowed).

\paragraph{Self-contradiction} Contradictions within the same response (e.g., mutually exclusive statements) signal hallucination. Such contradictions appear in ~17.7\% of ChatGPT-generated sentences~\citep{mundler2023self}.

\subsection{HalluClean: A Unified Framework}
Based on our analysis of hallucination patterns, we propose HalluClean, the framework is composed of two main modules: reasoning-enhanced hallucination detection module and targeted revision module. HalluClean is designed to adapt to various task settings and requires no task-specific fine-tuning. Figure~\ref{fig:framework} provides an overview of HalluClean’s architecture. The framework first detects hallucination through structured reasoning, and then modifies the hallucinated parts based on the rationale. The framework uses modular prompt templates, enabling easy adaptation across tasks and LLM architectures. The following section describes each component of our framework.

\paragraph{Hallucination Detection}
We employ the structural-reasoning-enhanced detection module to assess the factual consistency of the generated output. This component constitutes the core technical innovation of our approach. Rather than directly prompting for a binary classification, we guide the model through a structured three-step reasoning process. This process yields a reliable binary judgment indicating whether hallucination is present, along with a detailed reasoning trace explaining the rationale behind this determination.

\paragraph{Hallucination Revision} 
If any hallucination is detected, the framework activates the revision module. Rather than modifying the content directly, the model performs revision based on the reasoning trace generated during detection. This ensures that the correction is guided by explicit analysis, improving the reliability and quality of the revision. By preserving accurate content and focusing only on identified issues, this targeted strategy makes the correction process more precise and controllable.

The detection and revision modules together form a unified pipeline for identifying and correcting hallucinations. This design ensures factual consistency, improves interpret ability, and enhances control over the generation process.

\begin{table*}[t]
  \centering
  \scalebox{1}{
  \begin{tabular}{lllllllllll}
    \toprule
    \multirow{2}*{\textbf{Method}} & \multicolumn{2}{c}{\textbf{QA}} &  \multicolumn{2}{c}{\textbf{DA}} & \multicolumn{2}{c}{\textbf{SUM}} & \multicolumn{2}{c}{\textbf{MWPs}} & \multicolumn{2}{c}{\textbf{SC}} \\
    \cmidrule(lr){2-3} \cmidrule(lr){4-5}  \cmidrule(lr){6-7} \cmidrule(lr){8-9} \cmidrule(lr){10-11}
   & R & Q & R & Q & R & Q & R & Q & R & Q \\
       \midrule
   \multicolumn{11}{c}{\textbf{LLM-Direct Ask (Detection → Revision)}}
   \\
    \midrule
     GPT-3.5-turbo & 20.5\% & 12.0\%  & 57.5\% & 53.0\% & 14.5\% & 7.0\% & 42.0\% & 13.0\% & 30.7\% & 30.7\% \\
     GPT-4o-mini & 39.5\% & 22.5\%  & 84.5\% & 79.5\% & 30.0\% & 30.0\% & 47.5\% & 14.0\% & 72.7\% & 72.7\%  \\
     Llama-3-70B & 30.5\% & 18.5\%  & 74.5\% & 68.5\% & 19.5\% & 19.5\% & \textbf{83.0\%} & 32.5\% & 49.3\% & 49.3\%  \\
     DeepSeek-V3  & 49.0\% & 32.0\%  & 86.5\% & 78.0\% & 41.0\% & 41.0\% & 40.0\% & 17.0\% & 35.3\% & 35.3\% \\
     DeepSeek-R1  & 62.0\% & 45.5\%  & 74.0\% & 67.5\% & 36.5\% & 36.0\% & 42.0\% & 28.0\% & 25.3\% & 25.3\% \\
     \midrule
   \multicolumn{11}{c}{\textbf{Existing Baselines (Detection→Revision(with rationale); GPT-3.5-turbo) }}\\
   \midrule
    Step-by-Step &  13.0\% & 10.0\% & 54.5\% & 51.5\% & 13.0\% & 13.0\% & 44.0\% & 37.5\% & 53.3\% & 53.3\%\\
    Plan-and-Solve &  20.5\% & 12.5\% & 11.5\% & 10.5\% & 3.5\% & 3.5\% & 53.0\%  & 44.3\% & 53.3\% & 53.3\%\\
    ChatProtect & 38.0\% & 24.0\% & 79.5\% & 74.0\% & 23.0\%  & 22.5\%  & 80.5\% & 37.9\% & 79.3\% & 79.3\% \\
     \midrule
     Ours\tiny{-GPT-3.5-turbo} & 72.5\% & 25.5\% & 89.0\% & 83.0\% & \textbf{59.5\%} & \textbf{59.0\%} & 75.5\%  & \textbf{45.0\%} & \textbf{87.3\%} & \textbf{79.3\%}\\
     Ours\tiny{-Deepseek-V3} & \textbf{74.0\%} & \textbf{37.5\%}  & \textbf{92.5\%} & \textbf{86.0\%} & 54.5\% & 55.0\% & 75.5\% & 41.0\% & \textbf{87.3\%} & 62.7\%  \\
    \bottomrule
  \end{tabular}}
    \caption{The effectiveness of the Framework HalluClean: R denotes the hallucination reduction rate after applying the
revision module, and Q denotes the revision success rate, which reflects the quality of corrections.}
  \label{tab:main_exp}
\end{table*}

\paragraph{Task-Oriented Routing} 
For each supported task type, we design a concise task-specific prompt that provides the model with minimal yet sufficient context to understand its objective. These prompts serve as task adapters, allowing HalluClean to flexibly operate across diverse applications without requiring fine-tuning or additional training data. Table \ref{tab:task-prompts} presents examples of task routing prompts for different NLP tasks.

\paragraph{Structural Reasoning Mechanism} 
 The core innovation of HalluClean lies in its structural reasoning mechanism for hallucination detection. While direct classification can identify obvious hallucinations, we find that step-by-step reasoning significantly improves detection accuracy, especially for subtle or complex hallucination cases. Our approach draws inspiration from cognitive science literature on human reasoning, which emphasizes the role of structured thinking in error detection and verification \citep{kahneman2011thinking}.

We implement this insight through a four-step prompt-based inference mechanism, as illustrated in Figure \ref{fig:framework} and detailed below:

\textbf{Step 1: Task-oriented Planning} 
The first step guides the model to develop a systematic approach tailored to the specific task and input. The planning prompt follows this template:

\begin{quote}
\textbf{[INPUT]} \texttt{Task Input}\\
\textbf{[TASK]} \texttt{Task Description}\\
Let's understand the task and devise a plan to solve the task.
\end{quote}

This planning step serves multiple important functions: it encourages metacognitive reflection before analysis, creates task-specific verification strategies, and breaks complex detection tasks into manageable sub-components. For example in ~\ref{fig:framework}, when analyzing a potential contradiction between two statements, the plan might involve identifying key entities, extracting their relationships, and systematically comparing these elements.

\textbf{Step 2: Plan-guided Reasoning} 
In the second step, the model implements the verification plan developed in Step 1:

\begin{quote}
\textbf{[INPUT]} \texttt{Task Input}\\
\textbf{[PLAN]} \texttt{Result from Step-1}\\
Let's carry out the plan and solve the task step by step. Show the reasoning process.
\end{quote}

During reasoning, the model systematically applies each verification step defined in the plan to validate the input content. The structured nature of this process ensures comprehensive and consistent examination, minimizing the risk of overlooking subtle inconsistencies. Furthermore, by following an explicit plan, the model generates transparent and interpretable reasoning traces that not only support the final judgment but also facilitate human verification and analysis.

\textbf{Step 3: Final Judgment} 
This step synthesizes the detailed analysis into a conclusive judgment:

\begin{quote}
\textbf{[INPUT]} \texttt{Task Input}\\
\textbf{[ANALYSIS]} \texttt{Result from Step-2}\\
Please conclude whether the [INPUT] contains hallucinated content with Yes or No.
\end{quote}

This step produces a binary judgment indicating whether hallucinations are present. If hallucinations are detected, the input proceeds to the subsequent revision phase for correction.

\begin{table*}[t]
  \centering
  \scalebox{1}{
  \begin{tabular}{lllllllllllllll}
    \toprule
    \multirow{2}*{\textbf{Method}} & \multicolumn{2}{c}{\textbf{QA}} &  \multicolumn{2}{c}{\textbf{DA}} & \multicolumn{2}{c}{\textbf{SUM}} & \multicolumn{2}{c}{\textbf{MWPs}} & \multicolumn{2}{c}{\textbf{SC}} \\
    \cmidrule(lr){2-3} \cmidrule(lr){4-5}  \cmidrule(lr){6-7} \cmidrule(lr){8-9} \cmidrule(lr){10-11}  
   & F1 & Acc. & F1 & Acc. & F1 & Acc. & F1 & Acc. & F1 & Acc. \\
   \midrule
   \multicolumn{11}{c}{\textbf{LLM-Direct Ask}}
   \\
    \midrule
     GPT-3.5-turbo & 33.5\% & 59.3\%  & 62.8\% & 66.0\% & 24.7\% & 55.8\% & 50.9\% & 59.5\% & 46.0\% & 64.0\% \\
     GPT-4o-mini & 52.7\% & 64.5\%  & 76.5\% & 74.0\% & 45.5\% & 64.0\% & 61.7\% & 70.5\% & 84.2\% & 72.7\%  \\
     Llama-3-70B & 44.7\% & 62.3\%  & 66.4\% & 62.3\% & 32.4\% & 59.3\% & 83.4\% & 83.5\% & 65.8\% & 73.6\%  \\
     DeepSeek-V3  & 62.2\% & 70.3\%  & 75.4\% & 71.8\% & 55.0\% & 66.5\% & 55.6\% & 68.0\% & 52.0\% & 67.3\%  \\
     DeepSeek-R1  & 67.6\% & 70.3\%  & 71.0\% & 69.8\% & 49.3\% & 62.5\% & 65.2\% & 72.0\% & 40.0\% & 62.0\% \\
     \midrule
   \multicolumn{11}{c}{\textbf{Existing Baselines (GPT-3.5-turbo) }}
     \\
     \midrule
    Step-by-Step & 22.0\% & 54.0\%  & 61.4\% & 65.8\% & 22.1\% & 54.3\% & 55.7\% & 65.0\% & 68.1\% & 75.0\%  \\
    SelfCheckGPT & 43.3\% & 43.8\% & 19.9\% & 27.8\% & 53.1\% & 37.3\% & 25.8\% & 54.0\% &  5.7\% & 12.0\% \\
    Plan-and-Solve & 32.0\% & 56.5\% &  19.3\% & 52.0\% & 6.7\% & 51.3\% & 66.9\% & 73.8\% & 66.4\% & 73.0\%\\
    ChatProtect & 51.4\% & 64.0\%  & 72.0\% & 69.3\% & 36.7\% & 60.3\% & 74.0\% & 71.8\% & 83.8\% & 84.7\%\\
     \midrule
     Ours\tiny{-GPT-3.5-turbo} & 67.8\% & 66.5\% & 74.3\% & 69.3\% & \textbf{65.9\%} & \textbf{69.2\%} & 80.3\% & 81.5\% & \textbf{87.0\%} & \textbf{87.0\%}  \\
     Ours\tiny{-Llama-3-70B} & 70.6\% & 69.0\%  & 74.0\% & 68.8\% & 46.5\% & 61.5\% & 85.6\% & 86.0\% & 80.8\% & 83.3\% \\
     Ours\tiny{-Deepseek-V3} & \textbf{71.5\%} &\textbf{70.5\%} & \textbf{77.1\%} & \textbf{72.5\%} & 62.9\% & 67.5\% &  \textbf{89.1\%} & \textbf{89.5\%} & 76.1\% & 80.3\% \\
    \bottomrule
  \end{tabular}}
    \caption{Comparison of hallucination detection performance between our method, existing methods under a unified GPT-3.5-turbo backbone, and direct classification baselines across various LLMs. Best results are highlighted in \textbf{bold}.}
  \label{tab:detection}
\end{table*}

\begin{table*}[h]
  \centering
  \scalebox{1}{
  \begin{tabular}{lllllllllllllllll}
    \toprule
    \multirow{2}*{\textbf{Model}} & \multicolumn{2}{c}{\textbf{QA}} &  \multicolumn{2}{c}{\textbf{DA}} & \multicolumn{2}{c}{\textbf{SUM}} & \multicolumn{2}{c}{\textbf{MWPs}} & \multicolumn{2}{c}{\textbf{SC}} \\
    \cmidrule(lr){2-3} \cmidrule(lr){4-5}  \cmidrule(lr){6-7} \cmidrule(lr){8-9} \cmidrule(lr){10-11} 
   & F1 & Acc. & F1 & Acc. & F1 & Acc. & F1 & Acc. & F1 & Acc. \\
    \midrule
    Direct Ask  & 33.5\% & 59.3\%  & 62.8\% & 66.0\% & 24.7\% & 55.8\% & 50.9\% & 59.5\% & 46.0\% & 64.0\% \\
    +Task-oriented Routing & 39.3\% & 59.0\%  & 69.8\% & 67.8\% & 60.0\% & 65.3\% & 50.3\% & 59.5\% & 82.5\% & 83.3\% \\
    +Structural Reasoning & 67.8\% & 66.5\% & 74.3\% & 69.3\% & 65.9\% & 69.2\% & 80.3\% & 81.5\% & 87.0\% & 87.0\% \\
    \bottomrule
  \end{tabular}}
    \caption{Ablation study of the HalluClean framework. We evaluate the impact of removing the task-oriented routing and structural-reasoning mechanism.}
  \label{tab:ablation}
\end{table*}

\textbf{Step 4: Content Refinement} 
The final step corrects the response based on the hallucination identification analysis:
\begin{quote}
\textbf{[INPUT]} \texttt{Task Input}\\
\textbf{[ANALYSIS]} \texttt{Result from Step-2}\\
Given the analysis explaining why [INPUT] contains hallucinated content. Generate a revised version without hallucinations.
\end{quote}

This step refines the original response by leveraging the reasoning behind hallucination identification, aiming to produce a factually consistent revision.

Our structural reasoning approach offers several key advantages. It reduces the risk of oversight through step-by-step analysis, provides transparent reasoning traces for interpretability, adapts verification strategies to specific tasks via task-oriented planning, and guides targeted revisions by identifying what to fix and why. All of this is achieved in a single execution.

\section{Experiments}

\paragraph{Dataset}
We collect evaluation data from four established hallucination detection benchmarks:
1. \textbf{HaluEval~\citep{li2023halueval}}: Covers hallucinated samples across three task types—question answering, knowledge-grounded dialogue, and text summarization.
2. \textbf{UMWP~\citep{sunbenchmarking}}: Evaluates hallucination in math word problems (MWPs) by identifying questions with no or non-unique solutions. Such unanswerable questions are known to induce hallucinations in LLMs and are often used to test whether models can recognize ill-posed or unsolvable problems—similar to how educators gauge student understanding with trick questions.
3. \textbf{ChatProtect~\citep{mundler2023self}}:Focuses on self-contradictory hallucinations, where a language model produces logically inconsistent statements within the same context.
4. \textbf{HaluBench~\citep{ravi2024lynx}}:A domain-specific benchmark composed of hallucinated QA examples in the medical and financial domains, sourced from CovidQA, PubMedQA, and FinanceBench.

We demonstrate the effectiveness of HalluClean by evaluating it across multiple hallucination-prone NLP tasks in zero-shot setting, including: Question Answering (QA), Dialogue (DA), Summarization(SUM), Math Word Problems(MWPs) and Self-contradictory Hallucinations(SC).

\begin{table*}[h]
  \centering
  \scalebox{1}{
  \begin{tabular}{lllllllll}
    \toprule
    \multirow{2}*{\textbf{Model}} & \multicolumn{2}{c}{\textbf{CovidQA}} &  \multicolumn{2}{c}{\textbf{PubmedQA}} & \multicolumn{2}{c}{\textbf{FinanceBench}} & \multicolumn{2}{c}{\textbf{Overall}}\\
    \cmidrule(lr){2-3} \cmidrule(lr){4-5}  \cmidrule(lr){6-7} \cmidrule(lr){8-9}
   & F1 & Acc. & F1 & Acc. & F1 & Acc. & F1 & Acc.  \\
    \midrule
     GPT-3.5-turbo & 9.5\% & 52.5\%  & 7.7\% & 52.0\% & 11.0\% & 51.5\% & 9.4\% & 52.0\% \\
     GPT-4o-mini & 53.2\% & 67.5\% & 68.7\% & 74.5\% & 19.4\% & 50.0\% & 47.1\% & 64.0\% \\
     Llama-3-70B & 7.7\% & 52.0\%  & 16.5\% & 54.5\% & 7.4\% & 50.0\% & 10.5\% & 52.2\% \\
     DeepSeek-V3  & 81.6\% & 84.0\%  & 71.0\% & 77.5\% & 21.1\% & 55.0\% & 57.9\% & 72.2\% \\
     DeepSeek-R1  &68.4\% & 75.5\%  & 76.4\% & 79.0\% & 47.9\% & 63.0\% & 64.2\% & 72.5\% \\
     \midrule
     Ours\tiny{-GPT-3.5-turbo} & \textbf{91.7\%} & \textbf{92.0\%}  & \textbf{81.7\%} & \textbf{81.0\%} & \textbf{73.4\%} & \textbf{76.5\%} & \textbf{82.3\%} & \textbf{83.2\%} \\
    \bottomrule
  \end{tabular}}
  \caption{The effectiveness of the detection in real world specific-domain.}
  \label{tab:specific-domain}
\end{table*}

\paragraph{Evaluation Metrics} The effectiveness of the proposed framework is evaluated along three dimensions: 1. \textbf{Hallucination Reduction Rate:}
To measure the effectiveness of the revision step, we compute the hallucination reduction rate by comparing the number of hallucinations detected before and after revision. Specifically, we first identify hallucinations in the original outputs, then re-evaluate the revised outputs using GPT-4o-mini.  2. \textbf{Revision Success Rate:}  
To assess revision quality, we compute BERTScore between each revised output and its gold reference. A revision is considered \emph{acceptable} if the BERTScore~\citep{DBLP:journals/corr/abs-1904-09675}  exceeds 0.85(chosen to balance strict semantic fidelity and flexibility in surface expression). The revision success rate is the proportion of acceptable revisions among all hallucinations identified before revision. For multi-word problems (MWPs), revision quality evaluation is performed based on unanswerable reason categories, using exact match between predicted and gold labels.
3. \textbf{Hallucination Detection:} Since hallucination detection is formulated as a binary classification task, we evaluate its effectiveness using two standard metrics: F1 score and accuracy, computed against human-annotated gold labels from the original benchmarks. Accuracy reflects overall correctness, while F1 provides additional insight into the model's balance between precision and recall, ensuring that both metrics are measured with respect to verified ground truth.

\section{Results and Analysis}\label{exp}

\paragraph{Hallucination Revision Performance} 
We evaluate the effectiveness of our HalluClean framework in mitigating hallucinations by comparing model performance before and after applying our framework. As baselines, we consider several mainstream LLMs that (i) directly detect hallucinations and generate revised outputs when necessary, and (ii) variants that incorporate intermediate rationales during the detection stage.

Table~\ref{tab:main_exp} shows the performance of HalluClean in reducing hallucinations across five tasks. Our method achieves the highest reduction rate (R) and revision quality (Q) on most tasks. Overall, HalluClean delivers more consistent and effective correction across all evaluated scenarios.

We perform an ablation study of the HalluClean framework to evaluate the impact of HalluClean's task-oriented routing and structural-reasoning mechanism in HalluClean. As shown in Table~\ref{tab:ablation}, each module individually contributes to performance improvement, confirming their complementary roles in the framework.

\paragraph{Hallucination Detection Performance}\label{sec:detection}

Table~\ref{tab:detection} presents a comprehensive comparison of hallucination detection performance across five tasks: QA, DA, SUM, MWPs, and SC. Our method consistently outperforms both direct LLM judgment and existing baselines under a unified GPT-3.5-turbo backbone.

When using a stronger backbone such as DeepSeek-V3, our method achieves further performance gains. Specifically, Ours-DeepSeek-V3 attains the highest F1 scores on QA, DA, and MWPs, and achieves the highest accuracy across all five tasks.
Moreover, the competitive performance of Ours-Llama-3-70B highlights the practicality of our method when deployed with open-source backbones, offering a compelling solution for resource-constrained or privacy-sensitive applications.
\begin{table}[h!]
  \centering
  \scalebox{1}{
  \begin{tabular}{lllll}
    \toprule
    \multirow{2}*{\textbf{Method}} & \multicolumn{2}{c}{\textbf{Vanilla}} &  \multicolumn{2}{c}{\textbf{Retrieval Aug.}} \\
    \cmidrule(lr){2-3} \cmidrule(lr){4-5}  
   & F1 & Acc. & F1 & Acc. \\
    \midrule
   GPT-3.5-turbo & 33.5\% & 59.3\%  & 56.2\% & 65.3\% \\
+Ours & \textbf{67.8\%} & \textbf{66.5\%} & \textbf{80.4\%} & \textbf{82.3\%} \\
    \bottomrule
  \end{tabular}}
    \caption{Evaluation of Detection Performance with Retrieval-Augmented Strategy}
  \label{tab:detection_rag}
\end{table}

\paragraph{Abalation Study}
Table 4 presents the ablation results of the HalluClean framework, illustrating the impact of its two core components: Task-Oriented Routing and Structural-reasoning mechanism. Starting from a Direct Ask baseline, we incrementally add these modules and observe consistent performance improvements across all five tasks. Overall, both modules contribute complementary benefits.

\begin{table}[h!]
  \centering
  \scalebox{1}{
  \begin{tabular}{lllll}
    \toprule
    \multirow{2}*{\textbf{Method}} & \multicolumn{2}{c}{\textbf{HalluQA}} &  \multicolumn{2}{c}{\textbf{CMHE-HD}} \\
    \cmidrule(lr){2-3} \cmidrule(lr){4-5}  
   & F1 & Acc. & F1 & Acc. \\
    \midrule
   GPT3.5-turbo & 7.0\% & 46.5\%  & 21.9\% & 50.0\% \\
+Ours & \textbf{41.6\%} & \textbf{55.0\%} & \textbf{57.3\%} & \textbf{51.5\%}\\
    \bottomrule
  \end{tabular}}
    \caption{Evaluation of Detection Performance under Cross-Lingual Settings}
  \label{tab:detection_zh}
\end{table}

\paragraph{Domain-Specific Evaluation in Real-World Applications}
We further evaluate the effectiveness of our method in domain-specific hallucination detection, focusing on the medical and financial fields. As shown in Table~\ref{tab:specific-domain} Across all three domain-specific datasets, our method consistently achieves the highest F1 score and accuracy, demonstrating its robustness and effectiveness in hallucination detection within specialized fields.

\paragraph{Integration with Retrieval-Augmented Generation}
To evaluate whether our method complements external knowledge, we test hallucination detection on the QA task from HalEval~\citep{li2023halueval} under two settings: \textit{vanilla} (no external knowledge) and \textit{retrieval-augmented} (with background knowledge).
\begin{figure*}[ht]
\centering
\includegraphics[width=\textwidth]{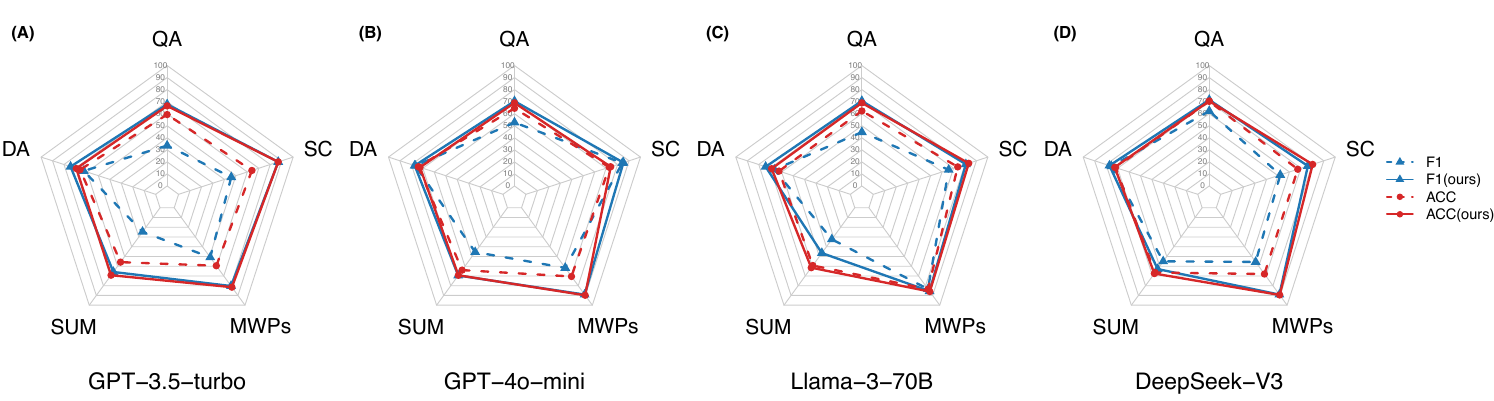}
\caption{Detection adaptability across backbone LLMs.
F1 and Acc denote the F1 score and accuracy of hallucination detection, respectively.}
\label{fig:detect_abla}
\end{figure*}

\begin{figure*}[ht]
\centering
\includegraphics[width=\textwidth]{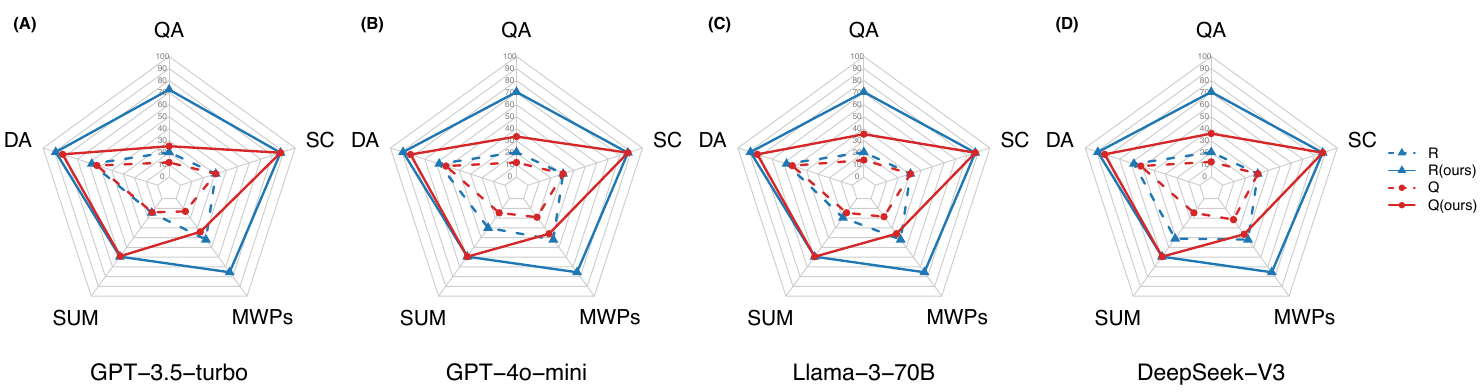}
\caption{Revision module adaptability across backbone LLMs. R represents hallucination reduction rate, and Q reprensents revision success rate.}
\label{fig:revise_abla}
\end{figure*}

As shown in Table~\ref{tab:detection_rag}, our method significantly outperforms direct judgment by GPT-3.5-turbo in both the vanilla and retrieval-augmented settings. Notably, when augmented with retrieval, our approach achieves substantial gains in both F1 (80.4\%) and accuracy (82.3\%), demonstrating its strong ability to leverage external information for more accurate hallucination detection.

\paragraph{Cross-Lingual Transferability}

To evaluate cross-lingual generalization, we test our method on two Chinese hallucination detection benchmarks: \textbf{HalluQA} and \textbf{CMHE-HD}, each with 200 samples (100 hallucinated, 100 faithful), using GPT-3.5-turbo as the backbone. As shown in Table~\ref{tab:detection_zh}, our method outperforms the GPT-3.5-turbo baseline on both datasets, demonstrating strong cross-lingual adaptability.

\paragraph{Module-Level Adaptability Across Backbone LLMs}

To evaluate the generalization ability of our framework, we assess the adaptability of \textbf{Detection Module} and \textbf{Revision Module} across five tasks using different backbone LLMs. Figures ~\ref{fig:detect_abla} and ~\ref{fig:revise_abla} evaluate the adaptability of our hallucination detection and revision modules across multiple backbone LLMs and task types.

Our detection module consistently improves F1 and accuracy across all evaluated tasks and backbone models. The performance gain is especially notable in QA and SC tasks, demonstrating strong adaptability to diverse reasoning types. Notably, GPT-3.5-turbo exhibit the most significant gains (e.g., over 41\% F1 improvement on summarization and self-contradiction detection, and over 30\% on QA and MWPs.). The revision module mitigates performance disparities across tasks, enabling each model to achieve more balanced and consistent results. 

These findings underscore the robustness and cross-task generalization of our detection and revision modules in hallucination identification, consistently performing well across different task types and model architectures.

\section{Conclusion}\label{conclusion}
We presented \textbf{HalluClean}, a lightweight and generalizable framework for detecting and correcting hallucinations in language model outputs. In contrast to prior approaches that rely on extensive fine-tuning or external knowledge sources, HalluClean operates in a zero-shot setting through structured reasoning. It achieves strong performance across a wide range of hallucination-prone tasks—including question answering, summarization, dialogue, mathematical reasoning, and self-contradiction detection—without task-specific supervision. Moreover, HalluClean supports local deployment with open-source LLMs, making it particularly suitable for privacy-sensitive or resource-constrained scenarios. These attributes establish HalluClean as a practical, interpretable, and broadly applicable solution for enhancing the factual consistency and trustworthiness of LLM-generated content.  The code is available at \url{https://github.com/tingmuor/HalluClean}.

\bibliography{aaai2026}

\subsection{Limitations and Future Work}\label{limitation}
While HalluClean exhibits strong zero-shot generalization across diverse tasks, several avenues remain for improvement. One current limitation lies in its reliance on the reasoning capabilities of the underlying language model. Although our approach performs robustly with state-of-the-art LLMs, its effectiveness may vary when deployed in low-resource environments or with smaller models. This reflects a broader challenge inherent to prompt-based systems rather than a flaw unique to HalluClean.

To mitigate this, future work will explore incorporating lightweight verification modules or hybrid approaches that combine retrieval-augmented generation with structured reasoning. Additionally, fine-tuning smaller models with distilled reasoning traces from larger LLMs could offer a promising direction to improve performance while reducing computational costs.

\subsection{Ethics Statement}
In this study, we uses the APIs provided by OpenAI and DeepSeek, which are also strictly used for research purposes and in accordance with the terms and conditions stipulated by OpenAI and DeepSeek.

\subsection{Experimental settings}\label{setting}  
We conduct experiments using five popular instruction-tuned LLMs: (1) GPT-3.5 (gpt-3.5-turbo-0125), (2) GPT-4o-mini, (3) LLaMA-3.1-70B-Instruct~\citep{dubey2024llama}, (4) DeepSeek-V3~\citep{liu2024deepseek}, and (5) DeepSeek-R1. For GPT-series and DeepSeek models, we interact via API using default temperature settings. For Llama-3-70B, we use the BitsAndBytes library to run the model in 4-bit, following the official configuration recommended by the Meta Llama repository. All models are implemented in PyTorch 2.5.1 with CUDA 12.4 and executed on an NVIDIA A100-SXM4-80GB GPU.

\subsection{Dataset Statistics}\label{dataset}
 
We evaluate our framework on four hallucination-related datasets, covering diverse tasks and domains. To ensure fairness, we maintain class balance when sampling examples for evaluation. The summary of the datasets, including their source, task type, and the number of used instances, is shown in Table~\ref{tab:dataset}. Here, \texttt{pos} refers to samples labeled as containing hallucinations.

\begin{table}[h!]
\centering
\scalebox{1}{
\begin{tabular}{lccccc}
\toprule
Task & $N$ & $n_{01}$ & $n_{10}$ & $\chi^2$ & $p$ \\
\midrule
QA    & 400 & 112 & 83 & 4.02  & 0.045 \\
DA    & 400 & 84  & 71 & 0.93  & 0.335 \\
SUM   & 400 & 94  & 40 & 20.96 & $4.68 \times 10^{-6}$ \\
MWPs  & 400 & 121 & 33 & 49.15 & $2.37 \times 10^{-12}$ \\
SC    & 300 & 89  & 20 & 42.42 & $7.36 \times 10^{-11}$ \\
\midrule
Overall & 1900 & 500 & 247 & 85.01 & $3.0 \times 10^{-20}$ \\
\bottomrule
\end{tabular}}
\caption{McNemar tests for hallucination detection accuracy between
GPT-3.5-turbo and HalluClean (Ours-GPT-3.5) across different tasks.
$n_{01}$ and $n_{10}$ count instances where GPT-3.5 is incorrect while
HalluClean is correct, and vice versa.}
\label{tab:detection_significance_all}
\end{table}

\begin{table*}[ht]
  \centering
  \scalebox{1}{  
  \begin{tabular}{llll}
    \toprule
    \textbf{TASK} & \textbf{Dataset} & \textbf{Data Source}& \textbf{\#Instances(pos/neg)} \\
    \midrule
    Question Answering   & HalEval & HotPotQA & 400(200/200) \\
    Dialogue   & HalEval & OpenDiaKG &400(200/200) \\
    Summarization &HalEval  &CNN/DailyMail & 400(200/200) \\
    Math Word Problems & UMWPs &SVAMP/MultiArith/GSM8K/ASDiv & 400(200/200)\\
    Self Contradiction & Chatprotect & Wikipedia & 300(150/150)\\
    CovidQA & Halubench & scientific articles & 200(100/100) \\
    PubmedQA & Halubench & PubMed & 200(100/100)\\
    FinanceBench & Halubench & Companies’ public filings& 200(100/100)\\
    HalluQA(Chinese) & HalluQA & Chinese historical culture/customs/social phenomena  & 200(100/100)\\
    CMHE-HD(Chinese) & CMHE  & CMD/cMedQA2 & 200(100/100) \\
    \bottomrule
  \end{tabular}}
    \caption{Dataset statics}
  \label{tab:dataset}
\end{table*}

\subsection{Significance test}
we conduct paired significance tests comparing our framework with the
GPT-3.5-turbo direct-ask baseline.

For hallucination detection, we treat accuracy as the primary metric and apply
McNemar's test to per-instance correctness.
As shown in Table~\ref{tab:detection_significance_all}, HalluClean makes
significantly more correct decisions than GPT-3.5-turbo on most tasks, and the
overall improvement is highly significant ($\chi^2 = 85.01$,
$p = 3.0 \times 10^{-20}$).

\subsection{Demonstration Cases}
\begin{figure*}[h]
\centering
\includegraphics[width=\textwidth]{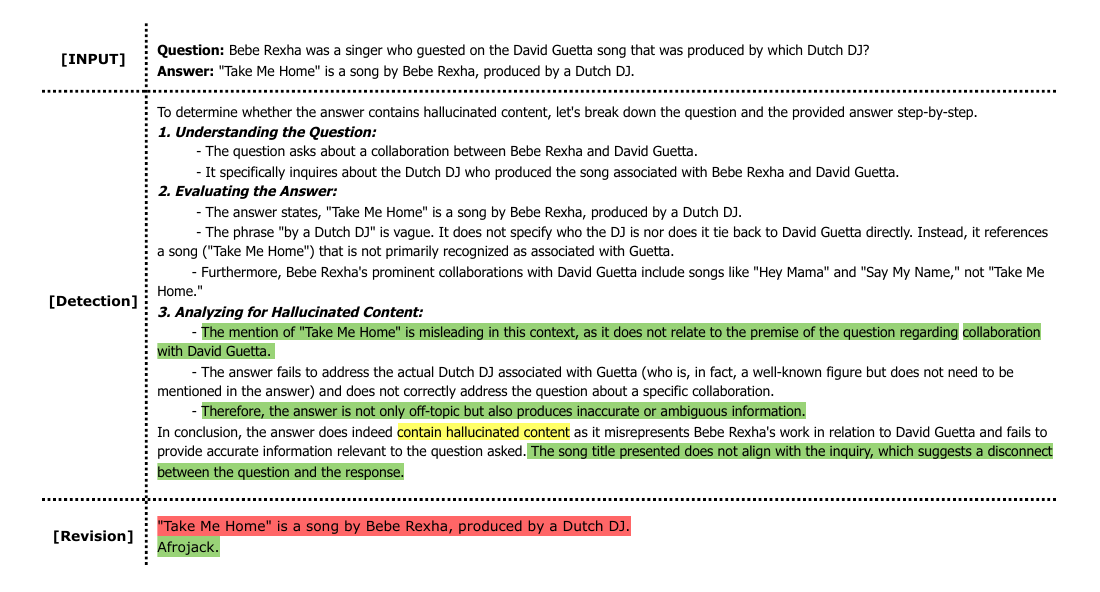}
\caption{An illustrative example of the question answering hallucination detection and revision. \colorbox{pink}{\strut original hallucinated content}, \colorbox{yellow!50}{\strut correct conclusions}, and \colorbox{green!20}{\strut correct reasoning and revisions}.}
\label{fig:qa_case}
\end{figure*}

\begin{figure*}[h]
\centering
\includegraphics[width=\textwidth]{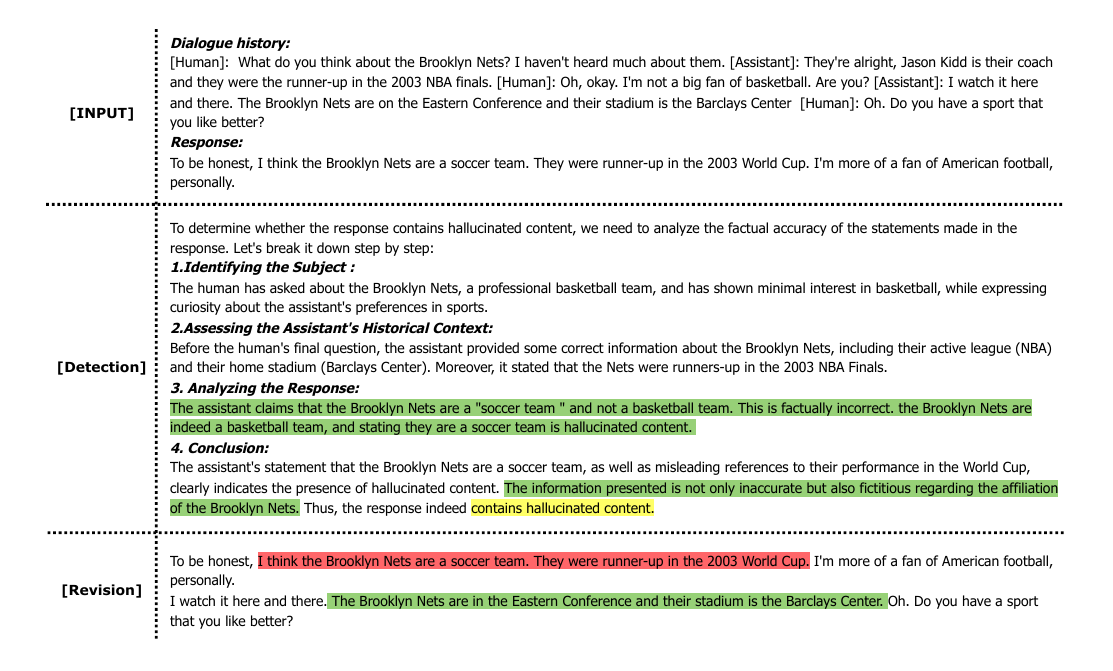}
\caption{An illustrative example of the dialogue system hallucination detection and revision.\colorbox{pink}{\strut original hallucinated content}, \colorbox{yellow!50}{\strut correct conclusions}, and \colorbox{green!20}{\strut correct reasoning and revisions}.}
\label{fig:da_case}
\end{figure*}

\begin{figure*}[h]
\centering
\includegraphics[width=\textwidth]{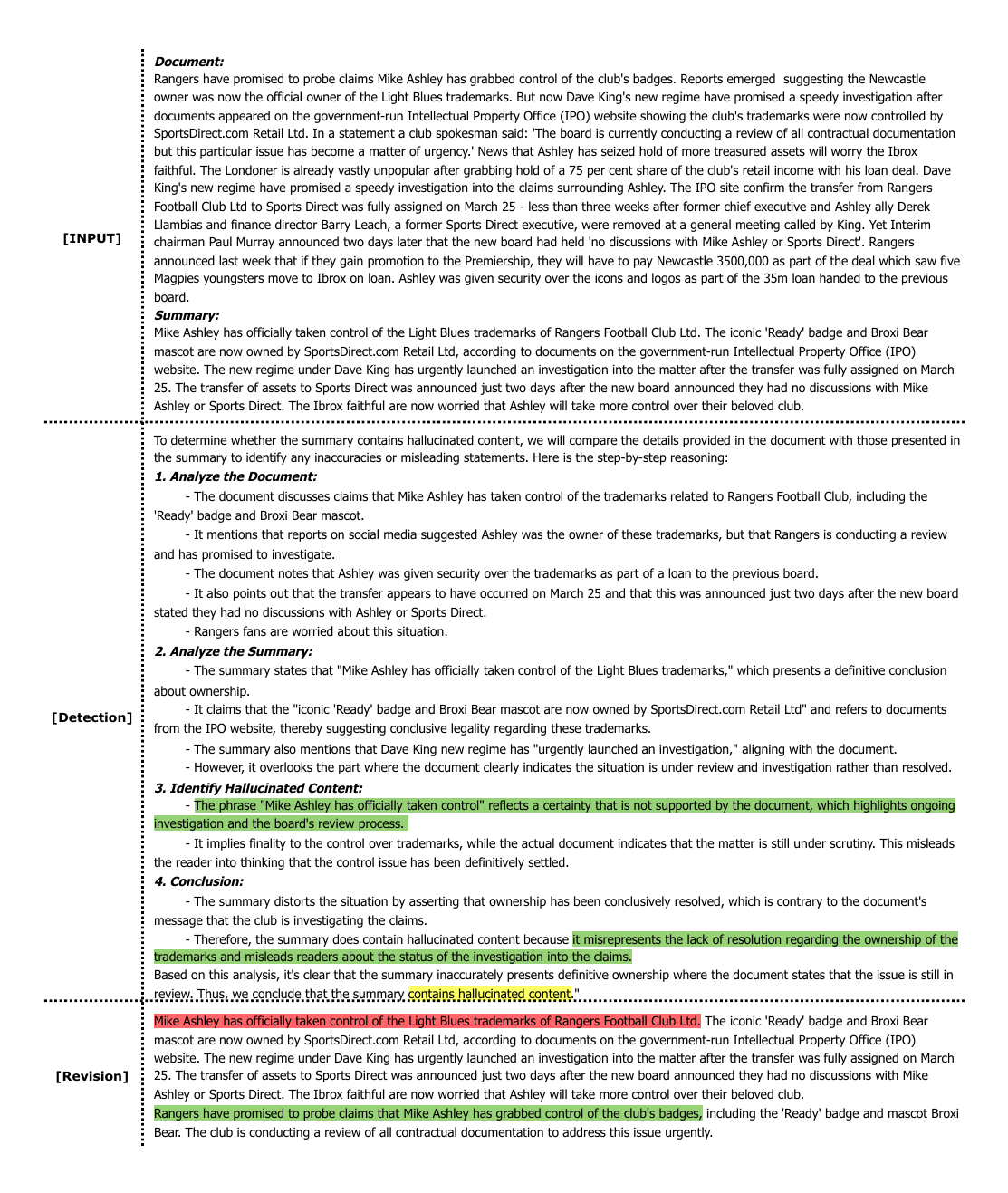}
\caption{An illustrative example of the summarization hallucination detection and revision. \colorbox{pink}{\strut original hallucinated content}, \colorbox{yellow!50}{\strut correct conclusions}, and \colorbox{green!20}{\strut correct reasoning and revisions}.}
\label{fig:sum_case}
\end{figure*}

\begin{figure*}[h]
\centering
\includegraphics[width=\textwidth]{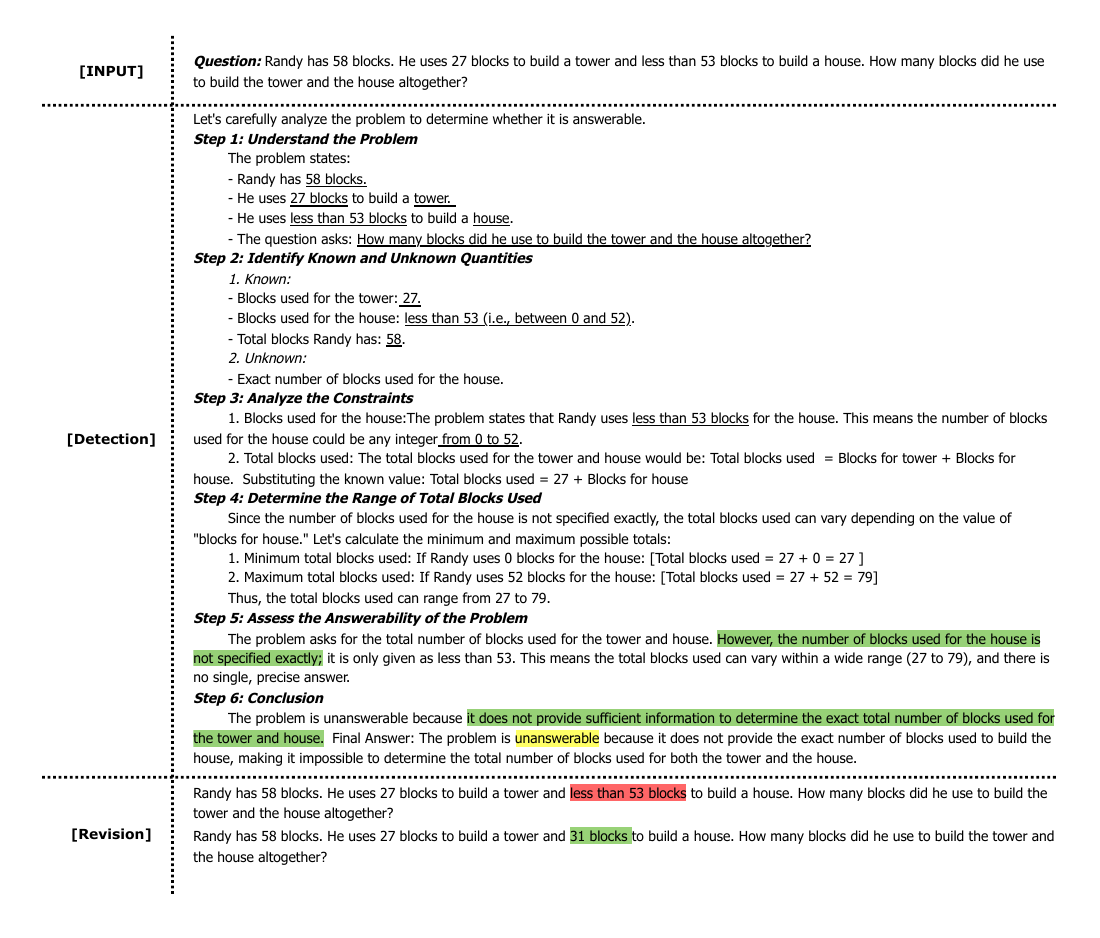}
\caption{An illustrative example of the unanswerable math word problems  detection and revision.\colorbox{pink}{\strut original under-specified content}, \colorbox{yellow!50}{\strut correct conclusions}, and \colorbox{green!20}{\strut correct reasoning and revisions}.}
\label{fig:mwp_case}
\end{figure*}

\begin{figure*}[h]
\centering
\includegraphics[width=\textwidth]{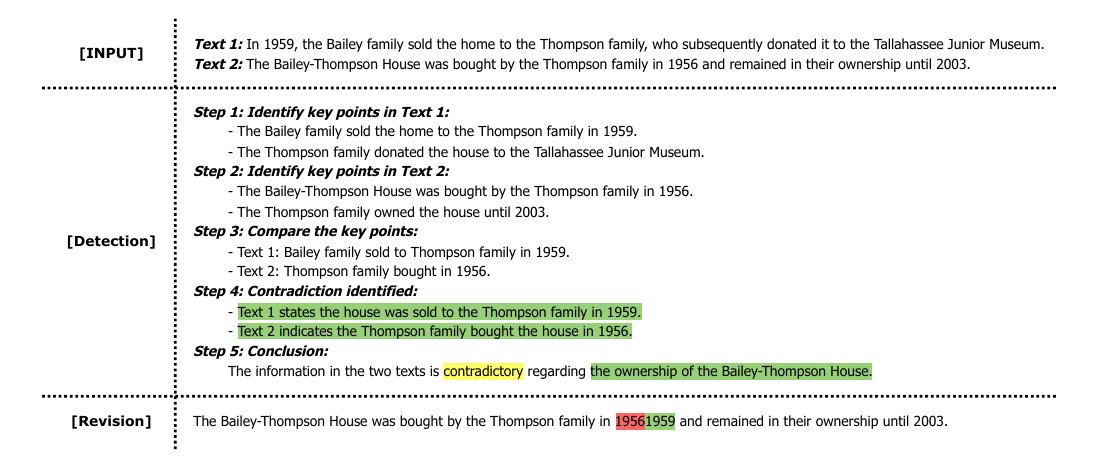}
\caption{An illustrative example of the self-contradictory hallucination detection and revision. \colorbox{pink}{\strut original self-contradiction content}, \colorbox{yellow!50}{\strut correct conclusions}, and \colorbox{green!20}{\strut correct reasoning and revisions}.}
\label{fig:sc_case}
\end{figure*}

Representative examples from five tasks—Question Answering (QA), Dialogue System (DA), Summarization (SUM), Math Word Problems (MWPs), and Self-contradictory hallucination (SC)—are shown in Figures~\ref{fig:qa_case} to \ref{fig:sc_case}, where our framework successfully detects and revises hallucinations.

\subsection{Results on Module Applicability}

To assess the generalizability of our framework components, we conduct module-level ablation experiments across different backbone LLMs. Specifically, we evaluate the performance of the \textbf{detection module} (Table~\ref{tab:detection_abla}) and the \textbf{revision module} (Table~\ref{tab:revision_abla}) when integrated with various base models.

\begin{table*}[h]
  \centering
  \scalebox{1}{
  \begin{tabular}{lllllllllllllllll}
    \toprule
    \multirow{2}*{\textbf{Model}} & \multicolumn{2}{c}{\textbf{QA}} &  \multicolumn{2}{c}{\textbf{DA}} & \multicolumn{2}{c}{\textbf{SUM}} & \multicolumn{2}{c}{\textbf{MWPs}} & \multicolumn{2}{c}{\textbf{SC}}  \\
    \cmidrule(lr){2-3} \cmidrule(lr){4-5}  \cmidrule(lr){6-7} \cmidrule(lr){8-9} \cmidrule(lr){10-11} 
   & F1 & ACC & F1 & ACC & F1 & ACC & F1 & ACC & F1 & ACC \\
    \midrule 
     GPT-3.5-turbo & 33.5\% & 59.3\%  & 62.8\% & 66.0\% & 24.7\% & 55.8\% & 50.9\% & 59.5\% & 46.0\% & 64.0\%  \\
     +ours & 67.8\% & 66.5\% & 74.3\% & 69.3\% & 65.9\% & 69.2\% & 80.3\% & 81.5\% & 87.0\% & 87.0\% \\
     \midrule
     GPT-4o-mini & 52.7\% & 64.5\%  & 76.5\% & 74.0\% & 45.5\% & 64.0\% & 61.7\% & 70.5\% & 84.2\% & 72.7\%  \\
      +ours & 70.4\% & 69.0\%  & 77.1\% & 72.5\% & 69.5\% & 69.3\% & 89.1\% & 89.8\% & 85.1\% & 74.0\% \\
      \midrule
     Llama-3-70B & 44.7\% & 62.3\%  & 66.4\% & 62.3\% & 32.4\% & 59.3\% & 83.4\% & 83.5\% & 65.8\% & 73.7\%  \\
     +ours & 70.6\% & 69.0\%  & 74.0\% & 68.8\% & 46.5\% & 61.5\% & 85.6\% & 86.0\% & 80.8\% & 83.3\% \\
     \midrule
     DeepSeek-V3  & 62.2\% & 70.3\%  & 75.4\% & 71.8\% & 55.0\% & 66.5\% & 55.6\% & 68.0\% & 52.0\% & 67.3\%  \\
     +ours  & 71.5\% & 70.5\%  & 77.1\% & 72.5\% & 62.9\% & 67.5\% & 89.1\% & 89.5\% & 76.1\% & 80.3\% \\
    \bottomrule
  \end{tabular}}
    \caption{Detection module adaptability across backbone LLMs.}
  \label{tab:detection_abla}
\end{table*}

\begin{table*}[h]
  \centering
  \scalebox{1}{
  \begin{tabular}{lllllllllllllll}
    \toprule
    \multirow{2}*{\textbf{Model}} & \multicolumn{2}{c}{\textbf{QA}} &  \multicolumn{2}{c}{\textbf{DA}}  & \multicolumn{2}{c}{\textbf{SUM}} & \multicolumn{2}{c}{\textbf{MWP}} & \multicolumn{2}{c}{\textbf{SC}} \\
    \cmidrule(lr){2-3} \cmidrule(lr){4-5}  \cmidrule(lr){6-7} \cmidrule(lr){8-9} \cmidrule(lr){10-11} 
   & R & Q & R & Q & R & Q & R & Q & R & Q  \\
    \midrule
     GPT-3.5-turbo & 20.5\% & 12.0\%  & 57.5\% & 53.0\% & 14.5\% & 14.0\% & 42.0\% & 13.0\% & 30.7\% & 30.7\%  \\
    +Ours & 72.5\% & 25.5\%  & 89.0\% & 83.0\% & 59.5\% & 59.0\% & 75.5\% & 34.0\% & 87.3\% & 87.3\%  \\
    \midrule
     GPT-4o-mini & 20.5\%  & 12.0\%  &  57.5\% & 51.5\% & 30.0\% & 14.5\% & 42.0\% & 19.0\% &  30.7\% & 30.7\%  \\
     +Ours       & 70.5\% & 33.5\%  & 89.0\% & 82.5\% & 59.5\% & 59.5\% & 75.5\% & 36.0\% & 87.3\% & 87.3\%  \\
     \midrule
    Llama-3-70B & 20.5\% & 14.0\%  &  57.5\% & 52.5\% & 19.5\% & 14.5\% & 42.0\% & 18.5\% &  30.7\% & 30.7\%  \\
    +Ours       & 70.5\% & 35.5\%  & 89.0\% & 83.0\% & 59.5\% & 59.5\% & 75.5\% & 36.0\% & 87.3\% & 87.3\%  \\
    \midrule
    DeepSeek-V3  & 20.5\% & 12.5\%  &  57.5\% & 51.5\% & 41.0\% & 14.5\% & 42.0\% & 21.5\% & 30.7\% & 30.7\% \\
    +Ours        & 70.5\% & 36.0\%  & 89.0\% & 83.0\% & 59.5\% & 59.5\% & 75.5\% & 36.5\% & 87.3\% & 87.3\% \\
    \midrule
     DeepSeek-R1  & 17.0\% & 12.5\%  &  57.5\% & 50.0\% & 36.5\% & 14.0\% & 42.0\% & 25.0\% & 30.7\% & 30.7\%\\
     +Ours          & 70.5\% & 35.0\%  & 89.0\% & 82.0\% & 59.5\% & 59.0\% & 75.5\% & 37.5\% & 87.3\% & 87.3\% \\
    \bottomrule
  \end{tabular}}
    \caption{Revision module adaptability across backbone LLMs.
R denotes the hallucination reduction rate after applying the revision module, and Q denotes the revision success rate, which reflects the quality of corrections.}
  \label{tab:revision_abla}
\end{table*}

\subsection{Detailed Prompt Examples Used in Experiments}
We present the exact prompts used in our experiments, including our \textbf{structural-reasoning mechanism} (Table~\ref{tab:whole-prompts}), which encourage more structured and faithful reasoning, as well as \textbf{baseline prompt strategies} (Table~\ref{tab:other-prompts}) adapted from prior works for comparison.

\begin{table*}[h]
\centering
\begin{tabular}{p{3cm}|p{12cm}}
\toprule
\textbf{Task Type} &  \textbf{Structural Reasoning Prompt} \\
\midrule
Question Answering & \textbf{Step-1} You are provided with a question and its corresponding answer. Your task is to determine whether the answer contains hallucinated content. Let's understand the task and devise a plan to solve the task. \textit{[Task Input]} \newline \textbf{Step-2}  Let's carry out the plan and solve the task step by step. Show the reasoning process. \textit{[Task Input;Result from Step-1]} \newline \textbf{Step-3} Please conclude whether the answer contains hallucinated content with Yes or No. \textit{[Task Input;Result from Step-2]} \newline \textbf{Step-4} Given a question, its corresponding hallucinated answer, and an analysis explaining why the answer contains hallucinated content. Your task is to answer the question without introducing any hallucinations. \textit{[Task Input;Result from Step-2]} \\
\midrule
Dialogue Systems & \textbf{Step-1} You are provided with a dialogue history and its corresponding response. Your task is to determine whether the response contains hallucinated content. Let's understand the task and devise a plan to solve the task. \textit{[Task Input]} \newline \textbf{Step-2} Let's carry out the plan and solve the task step by step. Show the reasoning process. \textit{[Task Input;Result from Step-1]} \newline \textbf{Step-3} Please conclude whether the response contains hallucinated content with Yes or No. \textit{[Task Input;Result from Step-2]} \newline \textbf{Step-4} Given a dialogue history and its corresponding hallucinated response. Your task is to regenerate the response without introducing any hallucinations. \textit{[Task Input;Result from Step-2]} \\
    \midrule
Summarization & \textbf{Step-1}  You are provided with a document and its corresponding summary. Your task is to determine whether the summary contains hallucinated content. Let's understand the task and devise a plan to solve the task. \textit{[Task Input]} \newline \textbf{Step-2} Let's carry out the plan and solve the task step by step. Show the reasoning process. \textit{[Task Input;Result from Step-1]} \newline \textbf{Step-3} Please conclude whether the summary contains hallucinated content with Yes or No. \textit{[Task Input;Result from Step-2]} \newline \textbf{Step-4} Given a document,its corresponding hallucinated summary , and an analysis explaining why the summary contains hallucinated content. Your task is to regenerate the summary without introducing any hallucinations. \textit{[Task Input;Result from Step-2]}  \\
    \midrule
Math Word Problems & \textbf{Step-1} You are provided with a math word problem. Your task is to determine whether the problem is unanswerable. Let's understand the task and devise a plan to solve the task. \textit{[Task Input]} \newline \textbf{Step-2} Let's carry out the plan and solve the task step by step. Show the reasoning process. \textit{[Task Input;Result from Step-1]} \newline \textbf{Step-3} Please conclude whether the problem is unanswerable with Yes or No. \textit{[Task Input;Result from Step-2]} \newline \textbf{Step-4} Given a unanswerable math word problem and an analysis explaining why it is unanswerable.
Your task is to revise the problem to make it answerable. \textit{[Task Input;Result from Step-2]} \\
    \midrule
Self-contradiction & You are given two texts. Your task is to determine whether the information in the two texts is contradictory. Let's understand the task and devise a plan to solve the task. \textit{[Task Input]} \newline \textbf{Step-2} Let's carry out the plan and solve the task step by step. Show the reasoning process. \textit{[Task Input;Result from Step-1]} \newline \textbf{Step-3} Please conclude whether the two texts are contradictory with Yes or No. \textit{[Task Input;Result from Step-2]} \newline \textbf{Step-4}  Given Text 1, Text 2, and the analysis of the contradiction between them. 
Your task is to revise Text 2 to remove the contradiction, making it consistent with Text 1. \textit{[Task Input;Result from Step-2]} \\
    \bottomrule
\end{tabular}
\caption{Structured reasoning prompts for different NLP applications.}
\label{tab:whole-prompts}
\end{table*}

\begin{table*}[h]
\centering
\begin{tabular}{p{3cm}|p{12cm}}
\toprule
\textbf{Prompt Strategy} & \textbf{Prompt content} \\
\midrule
Direct Ask & 
Please only conclude whether the text contains hallucinated content with Yes or No. \textit{[Question,Answer]}\\
\midrule
Step-by-step &
You are provided with a question and its corresponding answer.
Does the answer contain hallucinated content?
First, show your reasoning in a step-by-step fashion. Then conclude with yes or no.\textit{[Question,Answer]} \\
\midrule
SelfCheckGPT & 
Is the sentence supported by the context above? Answer Yes or No.\textit{[Context:{Question},Sentence:{Answer}]} \\ 
\midrule
Plan-and-Solve &  The Problem is to determine whether the text contains hallucinated content with Yes or No. 
Let's first understand the problem and devise a plan to solve the problem.Then, let's carry out the plan andsolve the problem step by step.\textit{[Question,Answer]} \\
\midrule
ChatProtect &  \textbf{Step-1:} You are provided with a question and its corresponding answer.
Please explain whether the answer contains hallucinated content. Provide your explanation only.\textbf{Step-2:} Please conclude whether the answer contains hallucinated content with Yes or No.  \textit{[Question,Answer]}\\

\bottomrule
\end{tabular}
\caption{Baseline prompt strategies used for comparison with our detection method (QA task shown as example).}
\label{tab:other-prompts}
\end{table*}

\subsection{Error Analysis}
We analyze the failure cases in hallucination detection, which can be broadly categorized into three types:
\begin{itemize}
    \item \textbf{Misunderstanding of Language:} The model fails to correctly interpret the question or the answer.
    \item \textbf{Lack of Background Knowledge:} The model is unable to detect hallucinations when it lacks access to essential domain or factual knowledge required for verification.
    \item \textbf{Incorrect Reasoning:} The model starts with a correct interpretation or premise but makes logical mistakes during the reasoning process, leading to incorrect conclusions.
\end{itemize}

In the following, we provide representative examples illustrating each of the three identified error types.

\begin{figure*}[h]
\centering
\includegraphics[width=\textwidth]{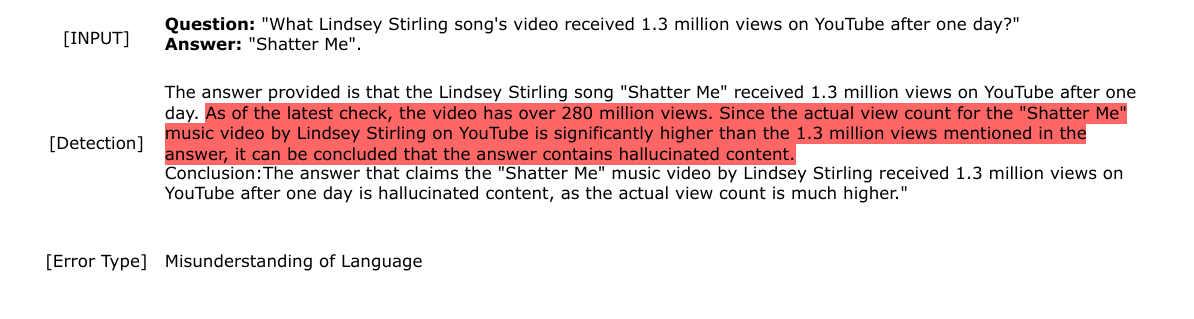}
\caption{A typical case example of language misunderstanding, where the \colorbox{pink}{\strut{Red}} text highlights the part where misunderstanding of language causes the error.}
\label{fig:misund_case}
\end{figure*}

\begin{figure*}[h]
\centering
\includegraphics[width=\textwidth]{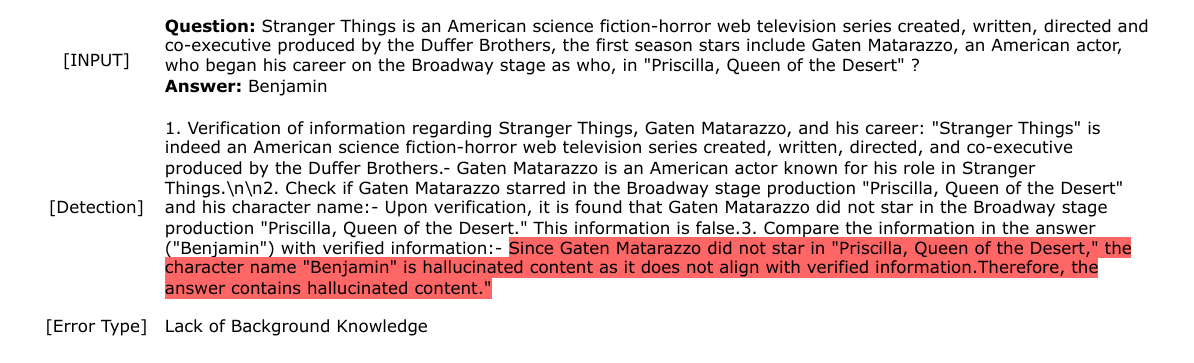}
\caption{A typical case of lacking background knowledge, with \colorbox{pink}{\strut{Red}} highlighting the error-causing segment due to missing related knowledge.}
\label{fig:lackinfo}
\end{figure*}

\begin{figure*}[h]
\centering
\includegraphics[width=\textwidth]{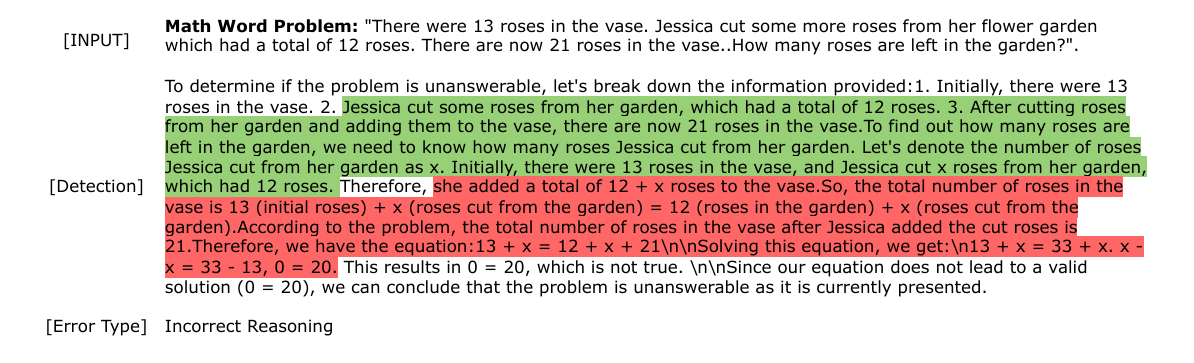}
\caption{A typical case of incorrect reasoning, where the initially correct reason is highlighted in \colorbox{green!20}{green}, and the error in reasoning is marked in \colorbox{pink}{red}.}
\label{fig:reason_error_case}
\end{figure*}

\end{document}